%% file: neurips_2025.tex
\documentclass{article}

    \PassOptionsToPackage{numbers, compress}{natbib}

    \usepackage[final]{neurips_2025}

\usepackage[utf8]{inputenc} 
\usepackage[T1]{fontenc}    
\usepackage{hyperref}       
\usepackage{url}            
\usepackage{booktabs}       
\usepackage{amsfonts}       
\usepackage{nicefrac}       
\usepackage{microtype}      
\usepackage{xcolor}         
\usepackage{colortbl}

\usepackage{graphicx}
\usepackage{multirow}
\usepackage{amsmath}
\usepackage{caption}
\usepackage{algorithm}
\usepackage{algorithmic}
\usepackage{graphicx}

\definecolor{HighlightColor}{gray}{0.9}

\title{\textbf{HunyuanVideo-Avatar: High-Fidelity Audio-Driven Human Animation for Multiple Characters}}

\author{%
    \raisebox{-0.2\height}{\includegraphics[width=0.8cm]{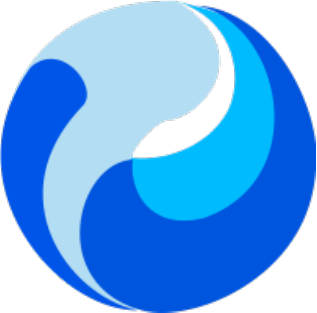}}
    \hspace{10pt}
    {\Large Tencent Hunyuan} \\  [0.5cm]
    \vspace{0.01cm}
{
\renewcommand{\arraystretch}{1.5}
\begin{tabular}{>{\centering\arraybackslash}p{1cm} p{10cm}}
\textbf{Page}: & \url{https://hunyuanvideo-avatar.github.io} \\
\textbf{Model}:        & \url{https://huggingface.co/tencent/HunyuanVideo-Avatar} \\
\textbf{Code}:         & \url{https://github.com/Tencent-Hunyuan/HunyuanVideo-Avatar}
\end{tabular}
}
}

\begin{document}

\maketitle

\input{sec/0_abstract}    
\input{sec/1_intro}
\input{sec/2_related_work}

\input{sec/3_method}
\input{sec/4_experiment}
\input{sec/5_conclusion}

\input{sec/7_supp}

\clearpage

\bibliography{neurips_2025}
\bibliographystyle{abbrvnat}

\end{document}

%% file: sec/0_abstract.tex
\begin{figure}[ht]
    \centering
    \includegraphics[width=1\textwidth]{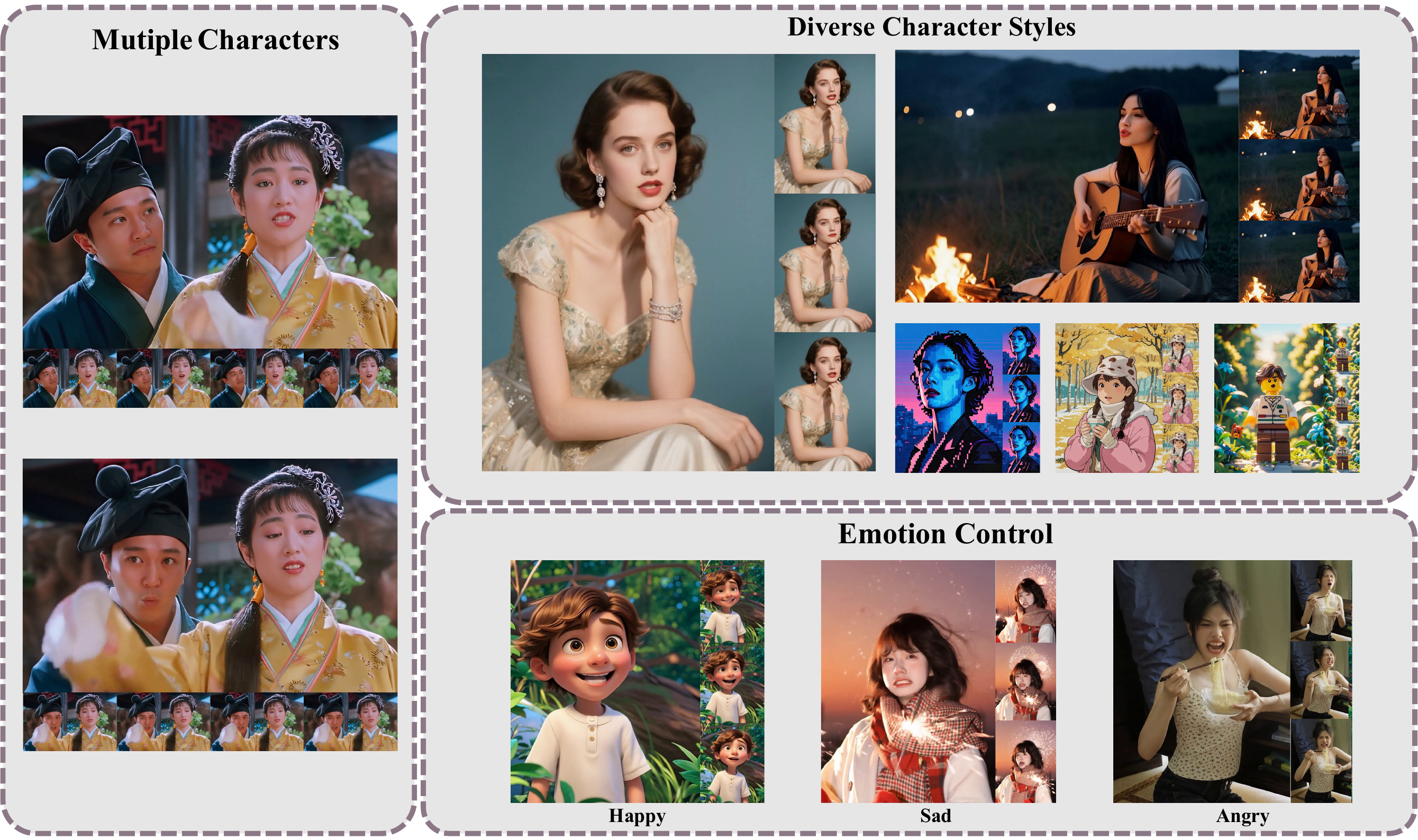}
    \caption{HunyuanVideo-Avatar can generate videos using a character image and audio as input. HunyuanVideo-Avatar enables the creation of multi-character, highly consistent, and dynamic human animations that accurately reflect the emotions expressed in the audio.}

    \label{fig:Data Construction Pipeline}
\end{figure}

\begin{abstract}
Recent years have witnessed significant progress in audio-driven human animation. However, critical challenges remain in (i) generating highly dynamic videos while preserving character consistency, (ii) achieving precise emotion alignment between characters and audio, and (iii) enabling multi-character audio-driven animation. To address these challenges, we propose HunyuanVideo-Avatar, a multimodal diffusion transformer (MM-DiT)-based model capable of simultaneously generating dynamic, emotion-controllable, and multi-character dialogue videos. Concretely, HunyuanVideo-Avatar introduces three key innovations: (i) A character image injection module is designed to replace the conventional addition-based character conditioning scheme, eliminating the inherent condition mismatch between training and inference. This ensures the dynamic motion and strong character consistency; (ii) An Audio Emotion Module (AEM) is introduced to extract and transfer the emotional cues from an emotion reference image to the target generated video, enabling fine-grained and accurate emotion style control; (iii) A Face-Aware Audio Adapter (FAA) is proposed to isolate the audio-driven character with latent-level face mask, enabling independent audio injection via cross-attention for multi-character scenarios. These innovations empower HunyuanVideo-Avatar to surpass state-of-the-art methods on benchmark datasets and a newly proposed wild dataset, generating realistic avatars in dynamic, immersive scenarios.
\end{abstract}

%% file: sec/1_intro.tex
\section{Introduction}
In recent years, Diffusion Transformers (DiT) have significantly advanced video generation. Among these developments, text-to-video and image-to-video techniques~\cite{bar2024lumiere,zhou2024allegro,svd,ayl,guo2023animatediff,zhou2022magicvideo,walt,wang2023modelscope,vdm,videogan,cvideogan,singer2022make,text2video,villegas2022phenaki,lin2025apt} have gained increasing attention due to their near-practical applicability. Audio-driven human animation, in particular, has experienced explosive growth, as it enables realistic human video synthesis with minimal input. Recent DiT-based approaches~\cite{zhang2023sadtalker, wang2024vexpress, echomimicv2, hallo3, lin2025omnihuman} have demonstrated superior performance in audio-driven generation compared to existing methods.

Current audio-driven human animation methods can be broadly categorized into two paradigms: portrait animation and full-body animation. Portrait animation methods~\cite{zhang2023sadtalker, wang2024vexpress, echomimicv2, hallo3, lin2025omnihuman} focus exclusively on facial movements while maintaining static or simplistic backgrounds. This narrow scope creates a fundamental disconnect between animated characters and their environments, often resulting in outputs that fail to meet practical expectations for immersive video content. Full-body animation methods~\cite{diffted, lin2024cyberhost, lin2025omnihuman} address this spatial limitation by extending motion to the full body. However, they face persistent challenges including unnatural character movements, misalignment between audio emotions and facial expressions, and an inability to drive multi-character scenes with audio. These limitations currently represent the most significant barrier to developing truly convincing audio-driven human animations.

Recent advances in audio-driven human animation have achieved significant progress, yet critical challenges persist in motion quality, character consistency, emotion alignment, and multi-character audio-driving. For instance, Hallo-3~\cite{hallo3}, a DiT-based portrait animation method, generates only facial movements while neglecting body motion. CyberHost~\cite{lin2024cyberhost} employs region attention and ReferenceNet to control facial and hand motions but often produces unrealistic movements in both the human body and background. OmniHuman-1~\cite{lin2025omnihuman} introduces a multimodal motion-conditioned hybrid training strategy to mitigate data scarcity issues, yet it still struggles with emotion-audio misalignment and multi-character scene generation. These limitations underscore the need for more robust solutions. To address these gaps, our work focuses on three key objectives: (i) improving dynamic expressiveness while preserving character identity, (ii) ensuring precise emotion synchronization between audio and video, and (iii) enabling realistic multi-character dialogue generation for real-world applications.

First, current audio-driven human animation methods typically rely on reference images during inference to enforce consistency between the generated video and the reference. However, this approach often leads to unnatural motion, as the model tends to replicate expressions and poses from the reference rather than generating dynamic, audio-aligned movements. To overcome this limitation, we propose a character image injection module, which transforms human image features into representations more amenable to model learning. By injecting these features along the channel dimension, we avoid the trade-off between dynamism and consistency that arises from direct latent space usage, ensuring coherence between training and inference. 

Second, we introduce an Audio Emotion Module (AEM) to align video characters' emotions with those conveyed in the audio. This module leverages reference images to guide emotion mapping, ensuring that facial expressions accurately reflect the audio's affective content, thereby improving realism in human animation.

Finally, to address the challenge of multi-character animation, we propose a Face-Aware Audio Adapter (FAA). This module applies a face mask to latent features extracted from the input, generating face-masked video latents that are then fused with audio information. Since the audio primarily influences the masked face region, we can independently drive different characters using distinct audio inputs, enabling realistic multi-character dialogue generation for cinematic applications. 

Extensive experiments demonstrate that our framework effectively drives multi-person scenarios with audio, significantly improving both dynamism and consistency in generated videos. The key modules of HunyuanVideo-Avatar are as follows:

\begin{itemize}
    \item  A character image injection module that resolves the dynamism-consistency trade-off caused by reference image usage, enhancing overall motion quality in foreground and background regions.
    \item An Audio Emotion Module (AEM) that aligns video characters' emotions with audio-driven affective cues, improving realism in facial expressions.
    \item  A Face-Aware Audio Adapter (FAA) that enables localized audio-driven animation for multiple characters by masking targeted face regions in the latent space, facilitating multi-character dialogue generation.
\end{itemize}

%% file: sec/2_related_work.tex
\section{Related work}
\noindent{\textbf{Audio-conditioned portrait animation.}} 
SadTalker~\cite{zhang2023sadtalker} generates 3D motion coefficients (including head pose and expression) from audio and implicitly modulates a novel 3D-aware face rendering technique, addressing issues of unnatural head movement, distorted expressions, and identity modification in existing talking head video generation, demonstrating superior performance in terms of motion and video quality. Hallo~\cite{xu2024hallo} proposes an innovative hierarchical audio-driven visual synthesis approach based on diffusion models, which seamlessly integrates generative models, denoisers, temporal alignment techniques, and a reference network to achieve precise synchronization between audio inputs and visual outputs, thereby enhancing the performance of portrait image animation in terms of image quality, lip-sync accuracy, and motion diversity. V-Express~\cite{wang2024vexpress} balances strong and weak control signals through progressive drop operations, enabling effective use of weak signals like audio in portrait video generation while considering pose, input image, and audio. Experiments validate its effectiveness. EchoMimic~\cite{echomimicv2} innovatively uses both audio and facial landmarks for training, addressing the instability and unnatural results of using audio or landmarks alone, enabling the generation of more natural portrait videos. Loopy~\cite{jiang2024loopy} learns natural motion and improves audio-portrait movement correlation through designed temporal modules and an audio-to-latents module, eliminating the need for manual motion templates to generate more realistic and high-quality videos.
Hallo3~\cite{hallo3} is designed with a Transformer-based identity reference network to ensure facial identity consistency, and explores speech audio conditions and motion frame mechanisms to enable the model's voice-driven capabilities.
In OmniHuman-1~\cite{lin2025omnihuman}, a multimodal motion condition hybrid training strategy is introduced, enabling the model to benefit from data augmentation with mixed conditions, thereby overcoming the challenges faced by previous end-to-end methods due to the scarcity of high-quality data.

\noindent{\textbf{Audio-conditioned full-body animation.}} DiffTED~\cite{diffted} is a novel one-shot audio-driven framework that leverages a diffusion model to generate synchronized and diverse talking head animations with natural co-speech gestures from a single image, using keypoint-guided Thin-Plate Spline motion modeling for temporally coherent and expressive video synthesis. CyberHost~\cite{lin2024cyberhost} includes two aspects: the Region Attention Module (RAM) maintains identity-independent latent features and combines identity-specific local visual features to enhance synthesis in key areas. Additionally, by introducing human prior-based conditions, it incorporates human structural priors into the model, reducing uncertainty in motion generation and improving video stability.

%% file: sec/3_method.tex
\section{Methods}
\begin{figure}[t]
    \centering
    \includegraphics[width=1\textwidth]{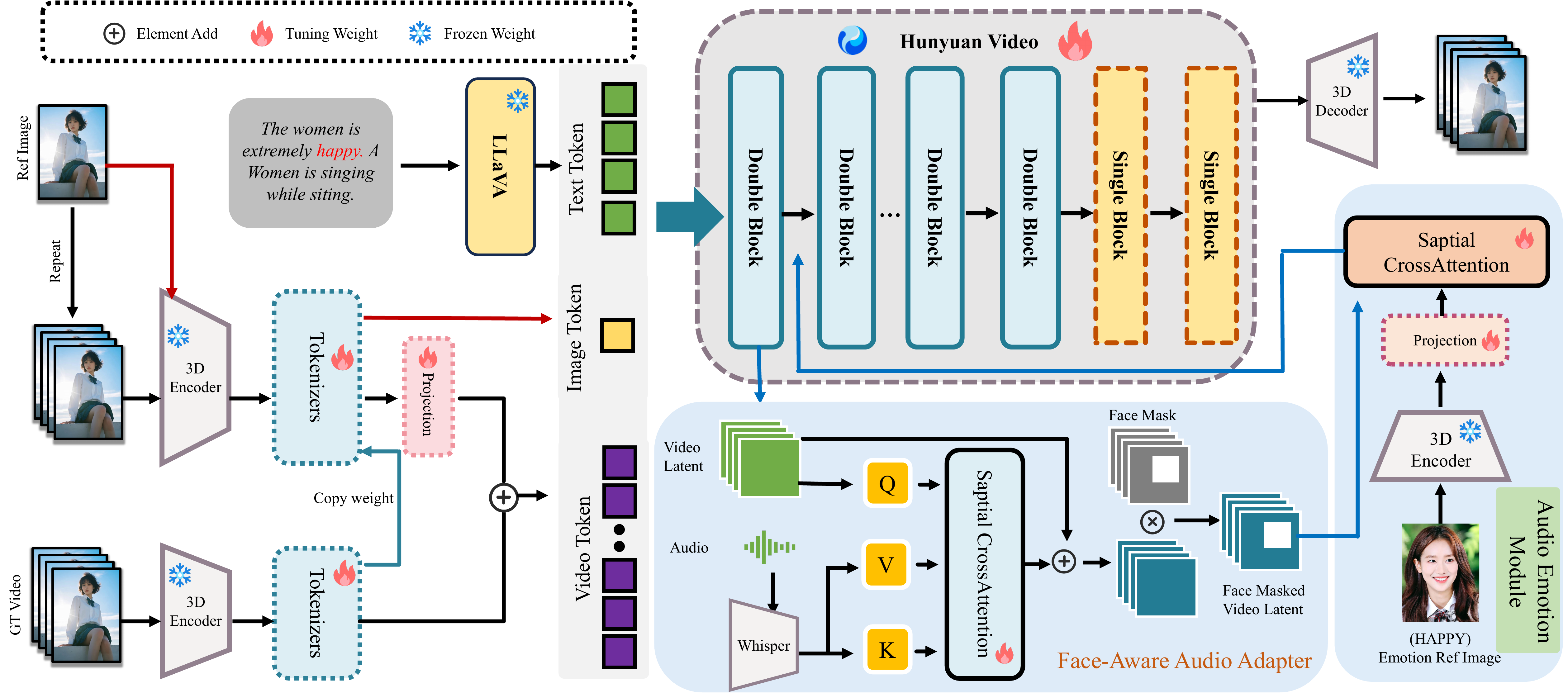}
    \caption{The framework of HunyuanVideo-Avatar. It consists of three parts: (1) Character Image Injection Module, which ensures high consistency of the character while maintaining high dynamics; (2) Audio Emotion Module, which aligns the character’s facial expressions in the video with the emotions in the audio; and (3) Face-aware Audio Adapter, which enables audio-driven multiple characters.}
    \label{fig:method_framework}
\end{figure}

Given a reference image, a driving audio, and a facial mask of the character, our method can generate talking videos of single or multiple characters based on the driving audio. The overall framework of our method is illustrated in the figure~\ref{fig:method_framework}. Specifically, we adopt HunyuanVideo~\cite{kong2024hunyuanvideo} as our backbone. It is a video generation model built upon the MM-DiT architecture.

In Section~\ref{sec:Preliminary}, we briefly introduced some preliminary knowledge.
In Section~\ref{sec:CIM}, we explore a character image injection module, which can maintain both character consistency and vividness. Then, in Section~\ref{sec:FAA}, we discuss how to apply an audio adapter to face region to enable multi-character audio-driven animation. In Section~\ref{sec:AMM} we discuss an emotion alignment module.  In Section~\ref{sec:long}, we briefly introduce the long video generation mechanism.

\subsection{Preliminaries}
\label{sec:Preliminary}
Firstly, we resize the target image to match the dimensions of the video frames. We then use the pretrained 3D VAE from HunyuanVideo to map the reference image $R$ from image space to the latent space, obtaining the reference image latent $v_R \in \mathbb{R}^{w \times h \times c}$, where $w$ and $h$ denote the width and height of the latent, and $c$ is the feature dimension. Similarly, we encode the noise video using the 3D VAE to obtain the video latent $v_{\text{noise}} \in \mathbb{R}^{f \times w \times h \times c}$, where $f$ is the the number of video frames. Next, we process $v_R $ and $v_{\text{noise}}$ with Tokenizer2 $K_2$ to obtain $t_R \in \mathbb{R}^{wh \times c} $ and $t_{\text{noise} } \in \mathbb{R}^{fwh \times c}$, respectively. We then replicate the reference image $T$ times (where $T$ is the original video length) to obtain $i_{\text{r}}$, and use the 3D VAE together with Tokenizer1 $K_1$ (initialized with the weights of Tokenizer2) to obtain $t_{\text{r}} \in \mathbb{R}^{fwh \times c}$. We add $t_{\text{r}}$ and $t_{\text{noise}}$ element-wise, and concatenate the result with $t_R$ along the token dimension to form the final input $p$, as shown below:

\begin{equation}
p = \text{TokenCat}\left( \left\{ K_1(t_{\text{r}}) + K_2(z_{\text{0}}) \right\}, t_R \right)
\end{equation}

Thanks to the strong temporal modeling prior of HunyuanVideo, identity information can be efficiently propagated along the time axis. Therefore, we assign 3D-RoPE~\citep{rope} positional encoding to the concatenated image latent. In the original HunyuanVideo, video latents are assigned 3D-RoPE along the time, width, and height axes; for a pixel at position $(l, i, j)$ (where $l$ is the frame index, $i$ is the width, and $j$ is the height), the RoPE is $RoPE(l, i, j)$. For the image latent, to enable effective broadcasting of identity information along the temporal sequence, we place it at the $-1$-th frame, i.e., before the first frame with time index $0$. Furthermore, inspired by Omnicontrol~\citep{ominicontrol} in controllable image generation, to prevent the model from simply copying and pasting the target image into the generated frames, we introduce a spatial shift for the image latent, as follows:

\begin{equation}
RoPE_{z_I}(l,i,j) = RoPE(-1, i+w, j+h).
\end{equation}

During the training process, we employ the Flow Matching~\cite{flowmatching} framework to optimize our video generation model. Specifically, we first extract the latent representation of the video, denoted as $z_1$, along with its corresponding reference image $R$. To introduce stochasticity, we sample a time step $t \in [0, 1]$ from a logit-normal distribution~\cite{esser2024scaling}. We then initialize the noise vector $z_0 \sim \mathcal{N}(0, I)$ from a standard Gaussian distribution. The training sample at time $t$, $z_t$, is constructed by linearly interpolating.

The model is trained to predict the velocity $u_t = \frac{dz_t}{dt}$ at each time step to guide the sample $z_t$ towards $z_1$. During optimization, the model outputs a predicted velocity $\lambda_t$, and the parameters are updated by minimizing the mean squared error between $\lambda_t$ and the ground-truth velocity $u_t$. The overall generation loss is defined as:

\begin{equation}
\mathcal{L}_{\text{generation}} = \mathbb{E}_{t, z_{0}, z_{1}} \left\| \lambda_{t} - u_{t} \right\|^{2}.
\end{equation}

This training strategy enables the model to effectively learn the underlying data distribution and generate high-quality, customized video content conditioned on the reference image.

\subsection{Character Image Injection Module} \label{sec:CIM}
\begin{figure}[t]
    \centering
    \includegraphics[width=1\textwidth]{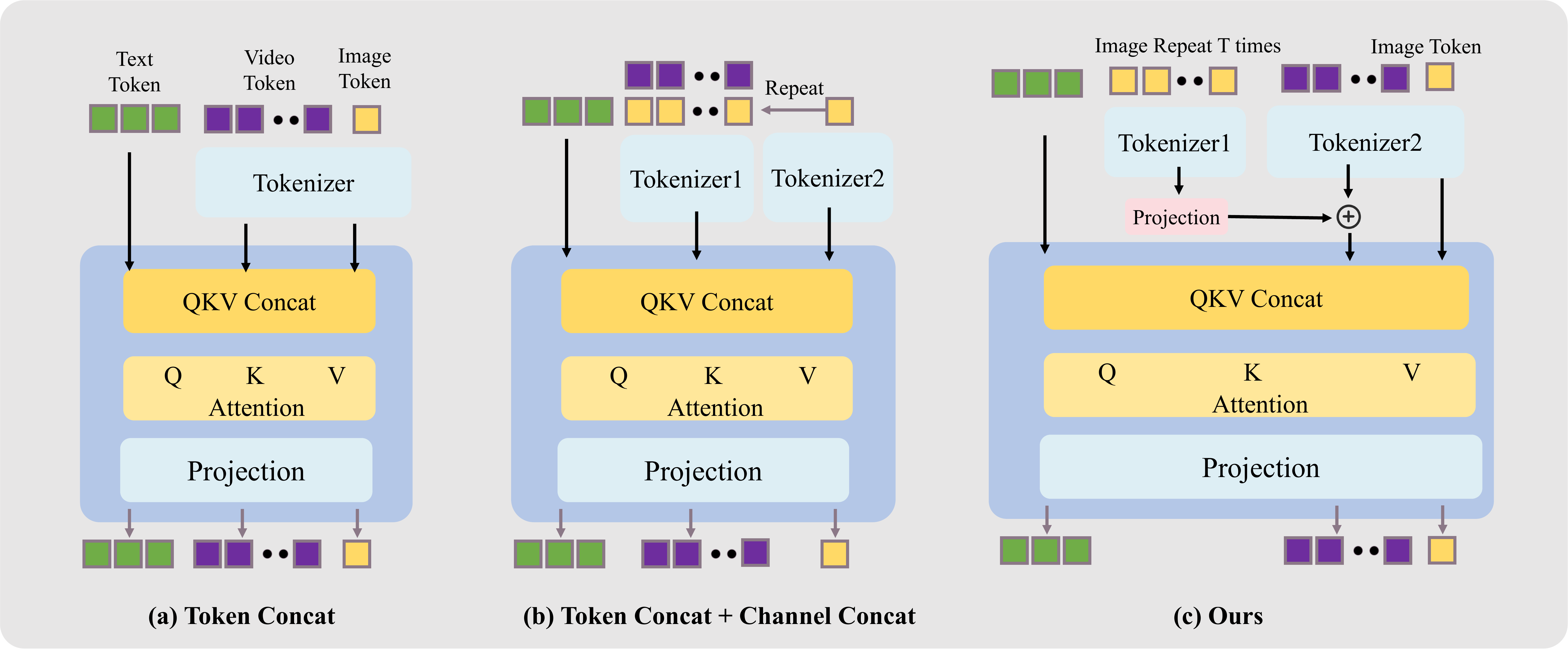}
    \caption{Three types of Character Image Injection Module.}
    \label{fig:CIM}
\end{figure}

In previous I2V methods, padding frames were often used for video inference. While this approach ensures good integrity and consistency of characters, backgrounds, and foregrounds, it also limits the motion dynamics of the generated video. Additionally, padding frames can lead to misalignment between the training and inference processes. Removing padding frames for video inference results in better motion dynamics but severely compromises character consistency and integrity. Therefore, we explored three character image injection mechanisms, as illustrated in Figure~\ref{fig:CIM}: (a) the reference image and video are processed through the same tokenizer, and the generated latents are concatenated in the token dimension; (b) the character image is first repeated T times (T represents the length of the video) and concatenated with the original video in the channel dimension, then fed into tokenizer1, while the character reference image is fed into tokenizer2, and both are concatenated in the token dimension before fed to the model; (c) the reference image is first repeated T times and fed into tokenizer2, then added directly to the video latent through a projection module composed of fully connected layers fed to the model. The mechanism (c) shows better results compared to mechanisms (a) and (b), as it improves the dynamics of motion while ensuring the consistency and integrity of characters, backgrounds, and foregrounds in the video, significantly enhancing video quality. For specific ablation comparisons experience, please refer to the experiment section. Since the backbone's tokenizer1 is specifically trained for video, we need to add an extra tokenizer2 to fit the image branch. The weights of this tokenizer are copied from the backbone's tokenizers, and we found that this approach accelerates model convergence.

\subsection{Face-aware Audio Adapter} \label{sec:FAA}
In terms of audio conditioning, we use Whisper~\cite{radford2023robust} for audio feature extraction, and for face masks, we employ the InsightFace~\cite{ren2023pbidr} method to detect the bounding box of the facial region. Given an audio-video sequence consisting of $n'$ frames, we extract audio features for each frame, yielding a feature tensor of shape $n' \times 10 \times d$, where 10 denotes the number of tokens per audio frame. The corresponding video latent representations are temporally compressed by a pretrained 3D VAE into $n$ frames, with $n = \left\lfloor \frac{n'}{4} \right\rfloor + 1$—where the additional 1 accounts for the initial, uncompressed frame, and 4 is the temporal compression ratio. Furthermore, to incorporate identity information, an identity image is concatenated at the beginning, resulting in a video latent of $n + 1$ frames.

To ensure temporal alignment between the audio features and the compressed video latent, we first pad the audio feature sequence prior to the initial frame, producing a total of $(n + 1) \times 4$ audio frames. We then aggregate every four consecutive audio frames into one, resulting in a temporally aligned audio feature tensor $g_A$ that matches the structure of the video latent representation:
\begin{equation}
    \begin{aligned}
        g_A=\text{Rearrange}(g_{A,0}): [b,(n+1)\times4,10,d]\rightarrow [b,(n+1),40,d]. \\
    \end{aligned}
\end{equation}
To ensure temporal alignment between the face mask and the compressed video latent, we set the face mask corresponding to the initial frame to 1, and also make it contain a total of $(n + 1) \times 4$ mask frames. This results in a mask $g_M$ that is both temporally and spatially aligned with the video latent.
With the temporally aligned audio features $g_A$, we introduce audio information into the video latent representation $y_t$ using a cross-attention mechanism. To prevent interference across different time steps, we adopt a \textbf{spatial cross-attention} strategy that performs audio injection separately for each time step. Specifically, each audio frame interacts only with the spatial tokens of its temporally aligned video frame, and cross-attention is applied independently at each temporal index. To this end, we decouple the temporal dimension from the spatial dimensions of the video latent and apply attention solely along the spatial axes:

\begin{equation}
    \begin{aligned}
        y'_{t,A}=\text{Rearrange}(y_t):&[b,(n+1)wh,d]\rightarrow [b,n+1,wh,d],\\
        y''_{t,A}=y'_{t,A}+&\alpha_A\times \text{CrossAttn}(g_A,y'_t) \times g_M,\\
        y_{t,A}=\text{Rearrange}(y''_{t,A}):&[b,n+1,wh,d]\rightarrow [b,(n+1)wh,d],
    \end{aligned}
\end{equation}

where $\alpha_A$ is a weight to control the influence of the audio feature.

\subsection{Audio Emotion Module} \label{sec:AMM}

To align the emotion conveyed in the audio with the character's facial expression, we compress the emotional reference image into features using a pretrained 3D VAE, and then inject these features into the \textbf{Double Block} of HunyuanVideo through an FC layer and \textbf{spatial cross-attention mechanism}. Specifically, the reference image features serve as the Key and Value, while the original video latent representation serves as the Query. This approach fuses information from the emotional reference image with the masked video latent $y_{t,A}$, enabling the model to better understand the relationship between audio emotion and facial expressions. To formalize this process, we first encode the emotional reference image $E_{\text{ref}} = \text{Encoder}(I_{\text{ref}})$, where $E_{\text{ref}}$ denotes the encoded feature of the emotional reference image $I_{\text{ref}}$. Next, to integrate these features into the video latent representation, we perform the following steps:
We first reshape the video latent $ y_{t,A} $ into temporal-spatial dimensions as $ y^{'}_{t,A} $, then apply an FC layer and spatial cross-attention to inject emotional features: $ y''_{t,A,E} $, and finally restore the original structure:

\begin{equation}
    \begin{aligned}
        y^{'}_{t,A} = \text{Rearrange}(y_{t,A}):& [b, (n+1)wh, d] \rightarrow [b, n+1, wh, d], \\
        y''_{t,A,E} = y'_{t,A} + &\gamma_E \times \text{CrossAttn}(\text{FC}(E_{\text{ref}}), y^{'}_{t,A}), \\
        y_{t,A,E} = \text{Rearrange}(y^{'}_{t,A,E}): &[b, n+1, wh, d] \rightarrow [b, (n+1)wh, d],
    \end{aligned}
\end{equation}

where $\gamma_E$ is a learnable scaling factor that controls the influence of the emotional reference features on the video latent.
Notably, we found that inserting this module into a \textbf{Single Block} does not allow the model to effectively learn emotional cues. In contrast, integrating it into a \textbf{Double Block} enables the model to better drive and express character emotions. This suggests that the Double Block architecture plays a crucial role in capturing and representing emotional details during complex emotion-to-expression mapping tasks.

\subsection{Long Video Generation}\label{sec:long}
The HunyuanVideo-13B model~\cite{kong2024hunyuanvideo} can only generate videos with 129 frames, which is often shorter than the audio length. To tackle the challenge of generating long videos, we use the Time-aware Position Shift Fusion method from Sonic~\cite{ji2024sonic}. We successfully adapt this method to the HunyuanVideo-13B model, which is based on the MM-DiT architecture, and achieve good results. This fusion strategy is simple yet effective, as it does not add any extra inference or training costs. It helps to reduce issues like jitter and abrupt transitions during video generation.

As shown in Algorithm~\ref{alg:shift-denoise}, at each timestep, the model takes a segment of the audio as input to predict the corresponding latent. It uses a starting offset to smoothly connect with the segment from the previous timestep, shifting forward by $\alpha$ steps each time. We set the offset $\alpha$ to 3-7 at each timestep, and our experiments show that this is an effective choice. This approach allows HunyuanVideo-Avatar to naturally bridge the context, enabling continuous video generation that follows the audio prompts.
\begin{algorithm}[ht!] 
\small
    \caption{Long Video Fusion} 
    \label{alg:shift-denoise} 
    \begin{algorithmic}[1] %
        \REQUIRE Audio embedding $v_a^{[0,l]}$ with length $l$, denoising steps $T$, initial noisy latent $z_T^{[0,l]}$, pretrained HunyuanVideo-Avatar model $\operatorname{HVA}(\cdot)$ for sequence length $f$, position-shift offset $\alpha < f < l$.\\ 
        \ENSURE Denoised latent $z_0^{[0,l]}$.
        \STATE Initialize accumulated shift offset $\alpha _{\beta}=0$. 
        \FOR{$t = T,\cdots,1$} \STATE \hfill// Denoising loop
        \STATE Initialize start point $s=\alpha _{\beta}$, end $e=s+f$, processed \\ length $n=0$. \hfill// Start from new position for each timestep.
        \WHILE{$n < l$} 
        \STATE \hfill // Sequence loop
        \STATE $z_{t-1}^{[s,e]} = \operatorname{HVA} (z_{t}^{[s,e]}, v_a^{[s,e]}, t)$ 
        \STATE $s \xleftarrow{} s + f$,  $e \xleftarrow{} e + f$, $n \xleftarrow{} n + f$. \hfill// Move to next clip non-overlapping
        \IF{$s > l$ \OR $e > l$}
        \STATE $s \xleftarrow{} s \% l $, $e \xleftarrow{} e \% l  $. \hfill// Padding circularly
        \ENDIF
        \ENDWHILE
        \STATE $\alpha _{\beta} \xleftarrow{} \alpha _{\beta} + \alpha $. \hfill// Accumulate shift offset
        \ENDFOR
        \RETURN Denoised latent $z_0^{[0,l]}$.
    \end{algorithmic}
\end{algorithm}

%% file: sec/4_experiment.tex
\section{Experiment}

\subsection{Experiment Settings}

\begin{figure}[ht]
    \centering
    \includegraphics[width=1\textwidth]{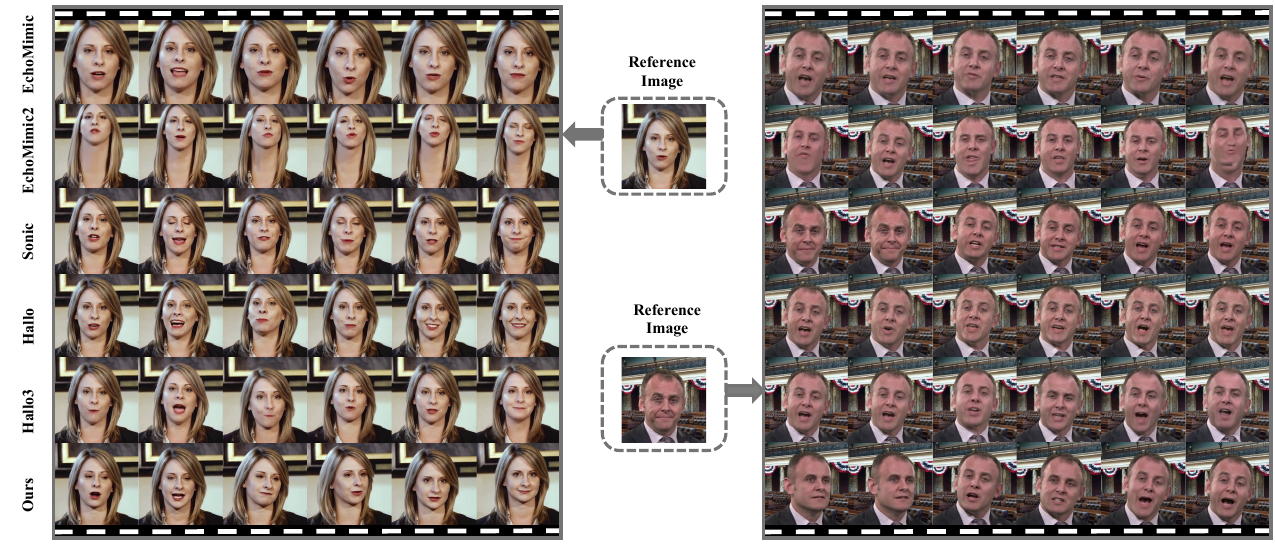}
    \caption{Qualitative comparison on the HTDF dataset.}
    \label{fig:qualitative_result1}
\end{figure}

\begin{figure}[ht]
    \centering
    \includegraphics[width=1\textwidth]{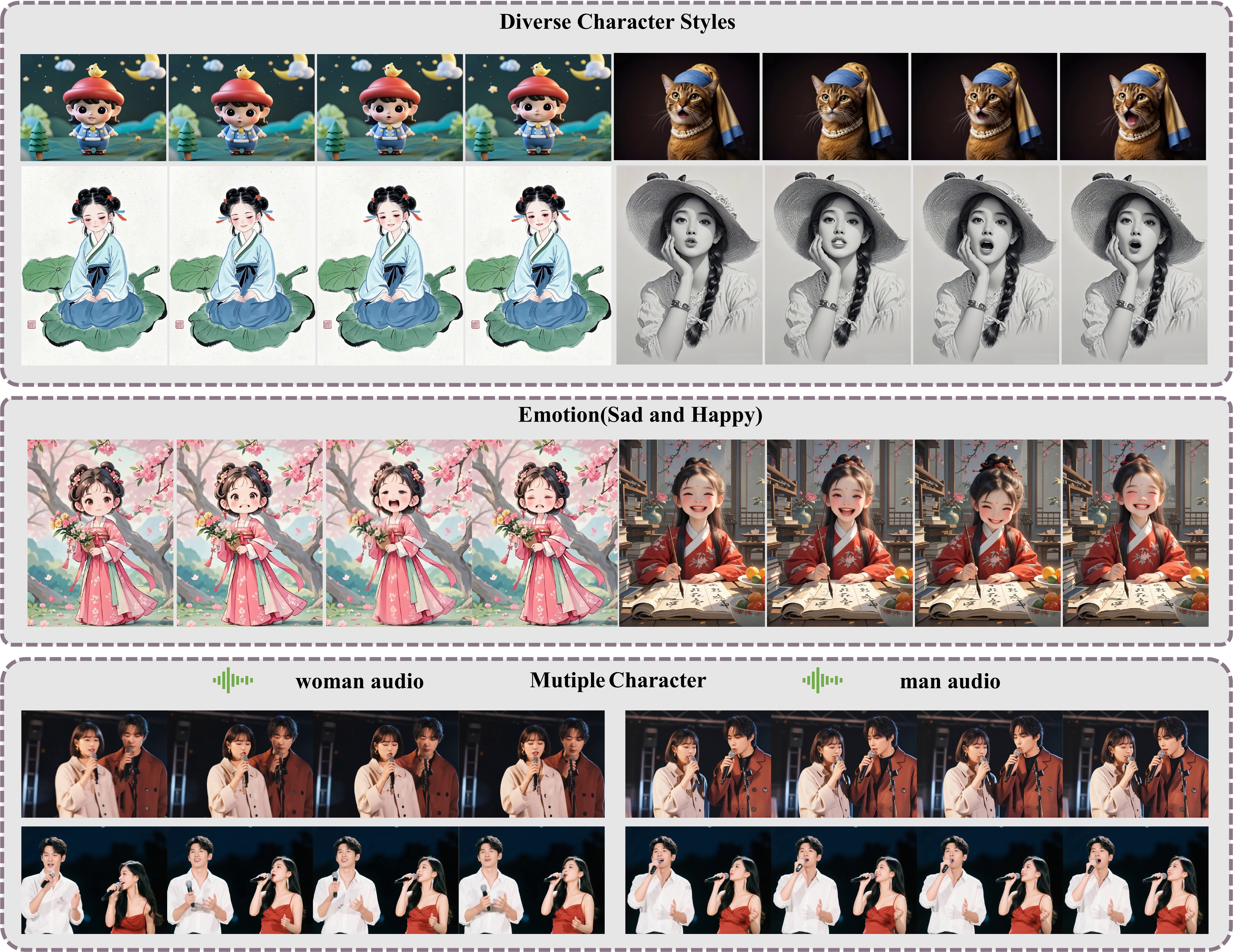}
    \caption{Visualization of videos generated by HunyuanVideo-Avatar on the wild dataset.}
    \label{fig:qualitative_result2}
\end{figure}

\begin{figure}[ht]
    \centering
    \includegraphics[width=1\textwidth]{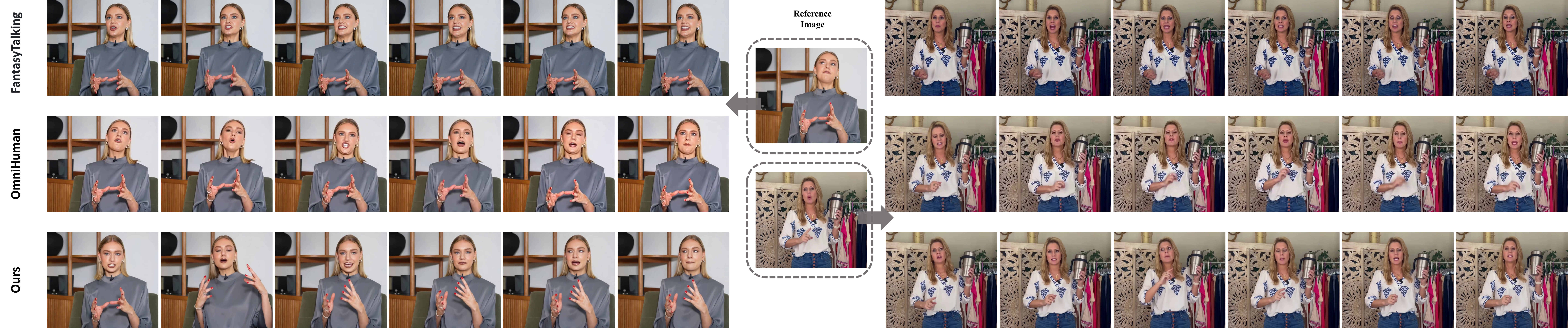}
    \caption{Qualitative comparison on the wild body dataset.}
    \label{fig:qualitative_result3}
\end{figure}

\paragraph{Implementation Details.}
We use HunyuanVideo-I2V ~\cite{kong2024hunyuanvideo} as the base model for HunyuanVideo-Avatar. The training process consists of two distinct stages. In the first stage, we train exclusively on audio-only data to establish fundamental audio-visual alignment. In the second stage, we implement a mixed training regime combining audio and image data in a 1:1.5 ratio to enhance motion stability. The resolution of the training data ranged from 704 $\times$ 704 to 704 $\times$ 1216. Throughout the training, we maintain fixed parameters for both LLaVA and 3D VAE while updating all other learnable parameters. We use 160 GPUs with 96GB of memory each, set the global batch size to 40, and the learning rate to 1e-5. 
\paragraph{Datasets.}
To obtain high-quality training data, we use LatentSync~\cite{li2024latentsync} to filter out audio-visual asynchronous data and employ tools such as Koala-36M~\cite{wang2024koala} to filter out data with low brightness or low aesthetics. Through this standardized data selection process, we obtain 500,000 samples with character audio for training, with a total duration of approximately 1,250 hours. During the testing stage, we select the publicly available portrait datasets CelebV-HQ~\cite{zhu2022celebv} (a dataset containing diverse scenes) and HDTF~\cite{zhang2021flow} (A dataset containing high-resolution videos and a larger number of subjects.) to evaluate the portrait animation capabilities of various methods. In addition, since there is currently no publicly available full-body animation test set, we construct our own full-body animation test set, which contains 250 videos covering 200 identities, involving different races, ages, genders, styles, and initial actions.
\paragraph{Evaluation Metrics and Compared Baselines.}
We use the Q-align~\cite{Q-align} visual language model (VLM) to evaluate video quality (IQA) and aesthetic metrics (ASE), and use FID~\cite{heusel2017gans} and FVD~\cite{unterthiner2019fvd} to assess the distance between generated videos and real videos. In addition, we use the smoothness metric from VBench~\cite{huang2024vbench} to evaluate video motion stability, and employ Sync-C~\cite{syncnet} to assess audio-visual synchronization. Apart from objective metrics, we also conducted a subjective evaluation with 30 users. The thirty users rated the generated results across four dimensions: lip synchronization(LS), Identity Preservation (IP), Full-body Naturalness(FBN), and Facial Naturalness(FCN).
To comprehensively assess the advancement of our method, we compared it with the current state-of-the-art audio-driven portrait animation methods, including Sonic~\cite{ji2024sonic}, 
EchoMimic~\cite{echomimicv2}, EchoMimic-V2~\cite{echomimicv2}, Hallo-3~\cite{hallo3}, and Omnihuman-1~\cite{lin2025omnihuman}. For audio-driven full-body animation, we first compared Hallo3~\cite{hallo3}, FantasyTalking~\cite{wang2025fantasytalking} and Omnihuman-1 on our proposed full-body animation test set. 

\subsection{Comparison with State-of-the-Art Methods}
\paragraph{Qualitative Results.}
We conducted qualitative comparisons with existing methods. For audio-driven portrait animation, we mainly compared our approach with Sonic, EchoMimic, EchoMimicV2, and Hallo-3 on the HDTF dataset, which primarily focuses on lip synchronization and facial expression accuracy. As shown in the figure~\ref{fig:qualitative_result1}, our method produces results with higher video quality, more natural and vivid facial expressions, and more aesthetically pleasing video effects on this dataset. 

For audio-driven full-body animation, The figure~\ref{fig:qualitative_result1} demonstrates the effectiveness of our model across various styles of characters, emotion control, and audio-driven multi-character scenarios, showcasing the validity of our approach. Then we mainly compared our method with Hallo3, FantasyTalking, and OmniHuman-1 on the wild full-body dataset. As illustrated in the figure~\ref{fig:qualitative_result3}, our method generates videos that exhibit more natural variations in the foreground, background, and character movements, while also achieving more accurate lip synchronization and better character consistency, resulting in higher overall video quality. These improvements are attributed to our method’s focus on the audio adapter module for audio-driven human animation and the character image injection module. Therefore, our approach is better suited to meet the demands of practical application scenarios. More comparative results and visual results are provided in the Appendix.

\paragraph{Quantitative Results.}
To thoroughly validate the superiority of our method in audio-driven portrait animation, we compared our approach with baseline methods on various evaluation metrics using the CelebV-HQ and HDTF test sets. As shown in the table~\ref{Tab1}, the results demonstrate that our method achieves the best performance in FID, FVD, IQA, ASE, and Sync-C, proving the effectiveness of our approach in audio-driven portrait animation and showcasing its capability in audio synchronization.

Meanwhile, to verify the superiority of our method in audio-driven full-body animation, we conducted a comparison with baseline methods on various evaluation metrics using our proposed test set. As shown in Table~\ref{Tab1}, the experimental results demonstrate that our method achieves the best performance on most evaluation metrics, proving its effectiveness in audio-driven portrait animation generation and showcasing its capability in audio-visual synchronization.

\begin{table}[t]
\centering
{
    \caption{\textbf{Quantitative comparisons with audio-driven portrait and full-body animation baselines.}}
    \label{Tab1}
    \centering
    \fontsize{10}{10}\selectfont
    \setlength{\tabcolsep}{5.5pt} 
    \renewcommand{\arraystretch}{1.3} 
    \begin{tabular}{cccccc}
    \toprule
    \multirow{2}{*}{Methods} & IQA $\uparrow$ & ASE$\uparrow$ & Sync-C$\uparrow$ & FID$\downarrow$ & FVD$\downarrow$  \\ 
    & \multicolumn{5}{c}{CelebV-HQ / HDTF}   \\
    \midrule
     Sonic & 3.60 / 3.86 & 2.43 / 2.41 & \textbf{5.58} / \textbf{5.81} & 49.28 / 40.50 & \textbf{415.04} / 413.94 \\ 
     EchoMimic & 3.39 / 3.64 & 2.25 / 2.23 & 3.41 / 4.07 & 46.74 / 45.38 & 450.98 / 410.05 \\ 
     EchoMimic-V2 & 2.75 / 3.36 & 1.97 / 2.15 & 4.11 / 3.39 & 46.37 / 39.73 & 862.24 / 487.75 \\ 
     Hallo-3 & 3.57 / 3.77 & 2.38 / 2.35 & 4.57 / 4.87 & 45.69 / 39.07 & 444.92 / 380.31 \\
     \rowcolor{HighlightColor}
     \textbf{Ours} & \textbf{3.70} / \textbf{3.99} & \textbf{2.52} / \textbf{2.54} & 4.92 / 5.30 & \textbf{43.42} / \textbf{38.01} & 445.02 / \textbf{358.71} \\
     \\ 
    \toprule
    Methods & \multicolumn{5}{c}{Full-body Test Set}  \\ 
    \midrule
    Hallo3 & 4.34 & 2.77 & 5.13 & 50.12 & 629.94  \\
    Fantasy& 4.63 & 3.02 & 3.68 & 58.24 & 677.67    \\
    OmniHuman-1 & 4.65 & 2.99 & 5.34  &  49.68 & 719.40  \\
    \rowcolor{HighlightColor}
    \textbf{Ours} & \textbf{4.66} & \textbf{3.03} & \textbf{5.56} & \textbf{49.38} & \textbf{650.54}  \\

     \bottomrule

    \end{tabular}
}
\vspace{0.1cm}

\end{table}

\paragraph{User Study.}
To further validate the effectiveness of our proposed method, we conducted a subjective evaluation on the wild full-body animation dataset. Each participant assessed four key dimensions: lip synchronization(LS), Identity Preservation (IP), Full-body Naturalness(FBN), and Facial Naturalness(FCN). A total of 30 participants rated each aspect on a scale from 1 to 5. As shown in the table~\ref{tab:user-study}, the results indicate that HunyuanVideo-Avatar outperforms existing baseline methods in the IP and LS evaluation dimensions, which is attributed to the enhancements brought by our Character Image Injection Module and Face-aware Audio Adapter. Since OmniHuman-1 is not open source and its online services include super-resolution operations, there is a natural visual advantage in subjective evaluations. In addition, our effect also inherits some inherent problems of Hunyuanvideo. Therefore, on FCN and FBN, our indicators have certain deficiencies compared with Omnihuman-1.

\begin{table}[t]
\centering
\begin{minipage}{0.45\textwidth}

    \centering
    \captionof{table}{\textbf{User Study results.}}
    \label{tab:user-study}
    \fontsize{9}{10}\selectfont
    \setlength{\tabcolsep}{4pt}
    \renewcommand{\arraystretch}{1.3} 
    \begin{tabular}{ccccccc}
    \toprule
        Methods & FCN $\uparrow$ & FBN $\uparrow$ & IP $\uparrow$ & LS $\uparrow$ \\
    \midrule
        Hallo3 & 2.91 & 2.59 & 4.28 & 3.61 \\
        Fantasy & 3.43 & 3.49 & 4.65 & 4.21 \\
        Omnihuman-1 & \textbf{4.11} & \textbf{4.18} & 4.79 & 4.61 \\
        \rowcolor{HighlightColor}
        Ours & 3.91 & 3.88 & \textbf{4.84} & \textbf{4.65} \\
    \bottomrule
    \end{tabular}
    
\end{minipage}
\begin{minipage}{0.5\textwidth}
    \centering
    \captionof{table}{\textbf{Ablation Study.}}
    \label{tab:ablation-CIM}
    \fontsize{9}{10}\selectfont
    \renewcommand{\arraystretch}{1.3} 
    \setlength{\tabcolsep}{4pt}
    \begin{tabular}{ccccc}
    \toprule
        Methods & VQ $\uparrow$ & MD $\uparrow$ & IP $\uparrow$ & LS $\uparrow$ \\
    \midrule
        Token & 2.86 & 3.58 & 4.402 & 4.239 \\
        Token + Channel & 4.4 & 2.336 & 4.576 & 4.43 \\
        \rowcolor{HighlightColor}
        Token + Add & \textbf{4.16} & \textbf{4.127} & \textbf{4.28} & \textbf{4.161} \\
    \bottomrule
    \end{tabular}
\end{minipage}
\end{table}

\subsection{Ablation Study And Discussion}
\begin{figure}[ht]
    \centering
    \includegraphics[width=1\textwidth]{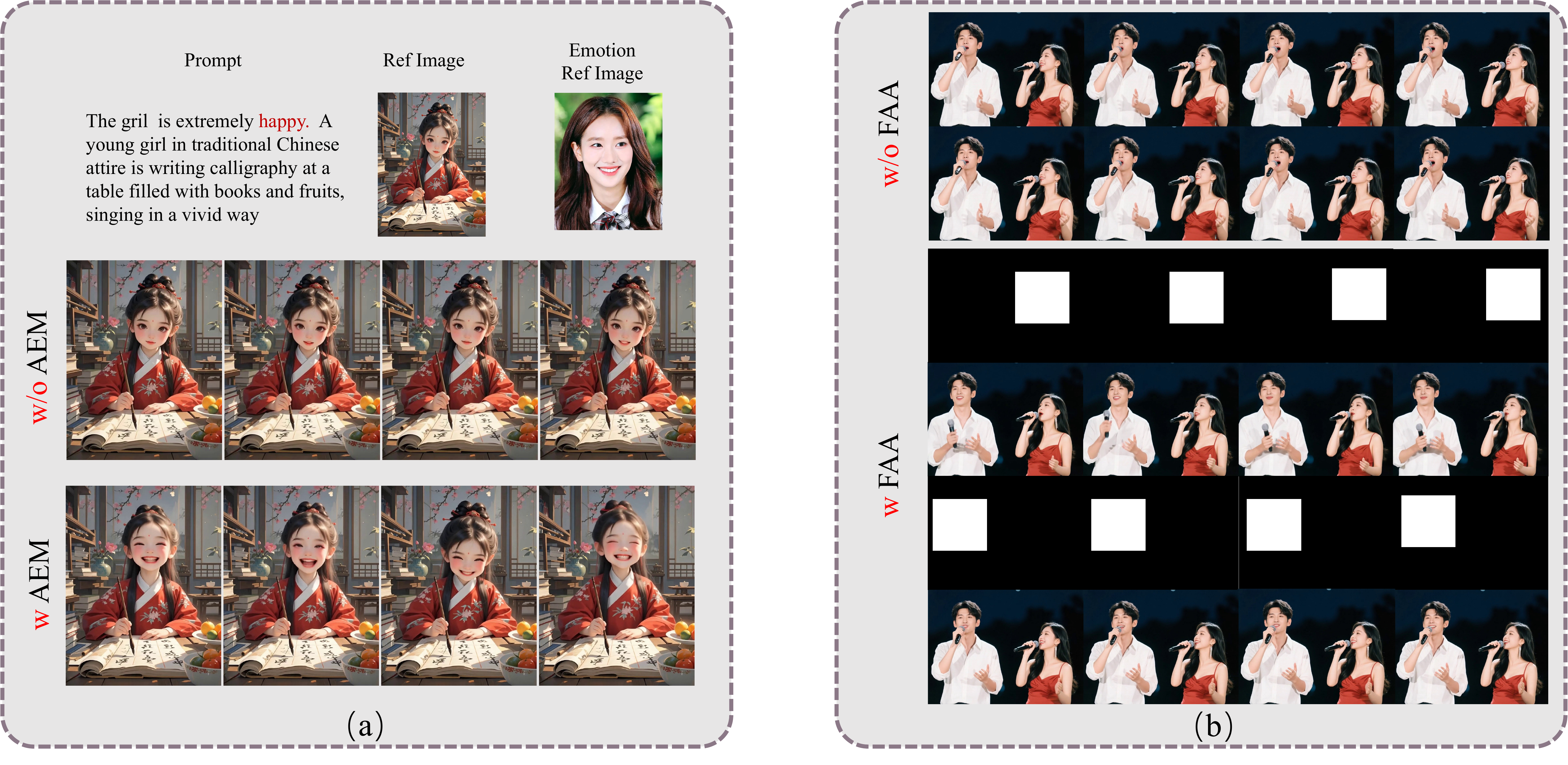}
    \caption{(a) Ablation on Audio Emotion Module. (b)Ablation on Face-Aware Audio Adapter. }
    \label{fig:qualitative_result2}
\end{figure}

\paragraph{Ablation on Character Image Injection Module. }
We subjectively evaluated three Character Image Injection Modules across four dimensions: Lip Synchronization (LS), Video Quality (VQ), Identity Preservation (IP), and Motion Diversity (MD). The results in the table~\ref{tab:ablation-CIM} indicate that our method performs better in terms of video dynamics and character consistency.

\paragraph{Ablation on Audio Emotion Module. } 
The figure~\ref{fig:qualitative_result2}\textbf{(a)} evaluates the impact of using the Audio Emotion Module on the facial emotions of video characters. The results show that if only text is used to guide the character's emotions without the Audio Emotion Module, the model cannot effectively understand or apply the emotions to the character's face. After injecting the emotion reference image into the model through the Audio Emotion Module, we find that the model can better transfer the emotional information from the reference image to the character's face, which helps us better align the emotions conveyed by the audio with the character's facial expressions.

\paragraph{Ablation on Face-Awared Audio Adapter. }
The figure~\ref{fig:qualitative_result2}\textbf{(b)} evaluates the impact of using the Face-Aware Audio Adapter for audio-driven animation of multiple characters. The results show that without restricting the region affected by audio using a Face Mask, both characters in the reference image are influenced by the audio information, causing the model to drive all characters with the audio. When the Face Mask is applied, we can see that the model drives only one specific character according to the mask, and as the Face Mask moves, the audio information is applied to the face of another character, thus enabling audio-driven multiple characters.

%% file: sec/5_conclusion.tex
\section{Conclusion}

In this paper, we propose HunyuanVideo-Avatar, an audio-driven human animation method that achieves both high character consistency and dynamic motion. We introduce a character image injection module resolves the inherent trade-off between dynamism and consistency by adaptively balancing these objectives, significantly enhancing the naturalness and diversity of generated videos. To ensure alignment between the audio's emotional tone and character expressions, we introduce the Audio Emotion Module which transfers affective cues from emotion reference images to the target animation. For multi-character scenarios, our method employs latent-space masking to localize audio-driven animation to specific face regions, enabling independent control of different characters through targeted mask modulation. Extensive qualitative and quantitative experiments demonstrate that HunyuanVideo-Avatar outperforms existing methods in video dynamism, subject consistency, lip-sync accuracy, audio-emotion-expression alignment, and multi-character scenarios.

%% file: sec/7_supp.tex
\section{Appendix}

\subsection{More Visualization Results}
\label{sec:Additional Results}

\begin{figure}[ht]
    \centering
    \includegraphics[width=1\textwidth]{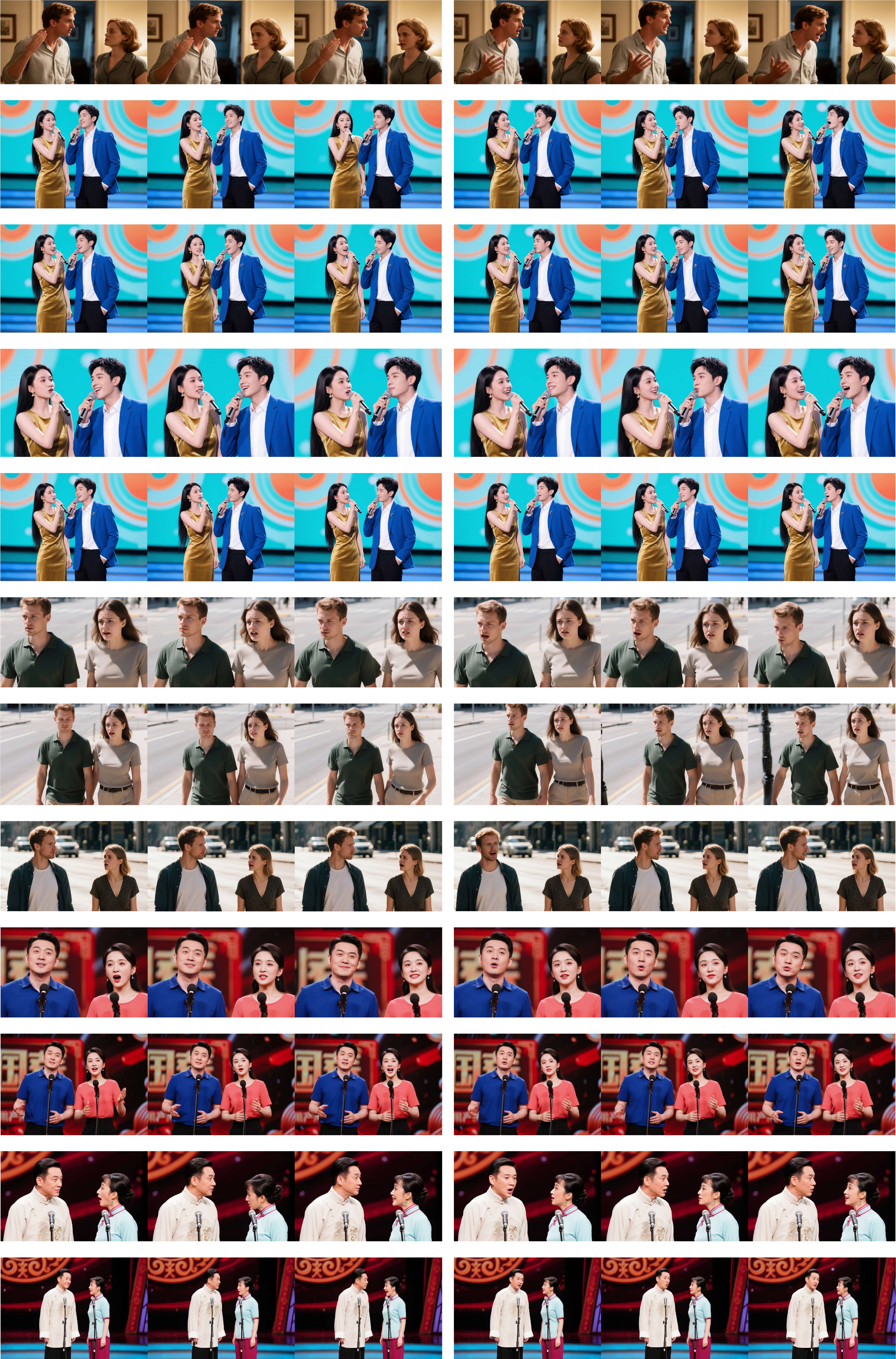}
    \caption{More visualizations on multi-character audio-driven human animation.}
    \label{fig:supp1}
\end{figure}

Figure~\ref{fig:supp1} shows the results of our method in multiple characters scenarios such as crosstalk, singing, and walking conversations, demonstrating the robustness of our model.  

\begin{figure}[ht]
    \centering
    \includegraphics[width=1\textwidth]{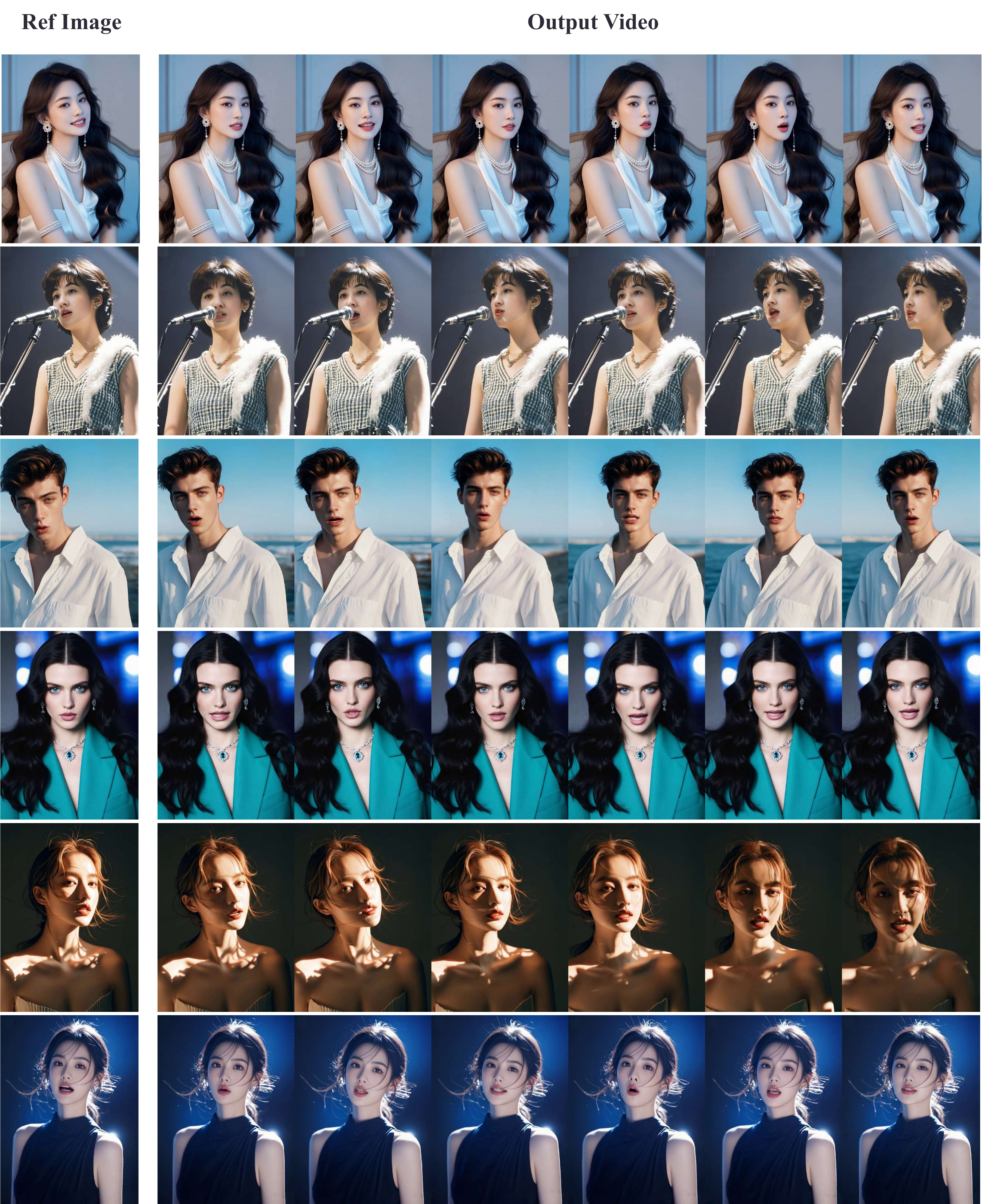}
    \caption{More visualizations on realistic scenarios.}
    \label{fig:supp2}
\end{figure}

Figure~\ref{fig:supp2} presents visualizations of realistic human images. From this scene, it can be seen that our model is able to maintain good character consistency while enhancing dynamics, further demonstrating the effectiveness of our character image injection module.

\begin{figure}[ht]
    \centering
    \includegraphics[width=1\textwidth]{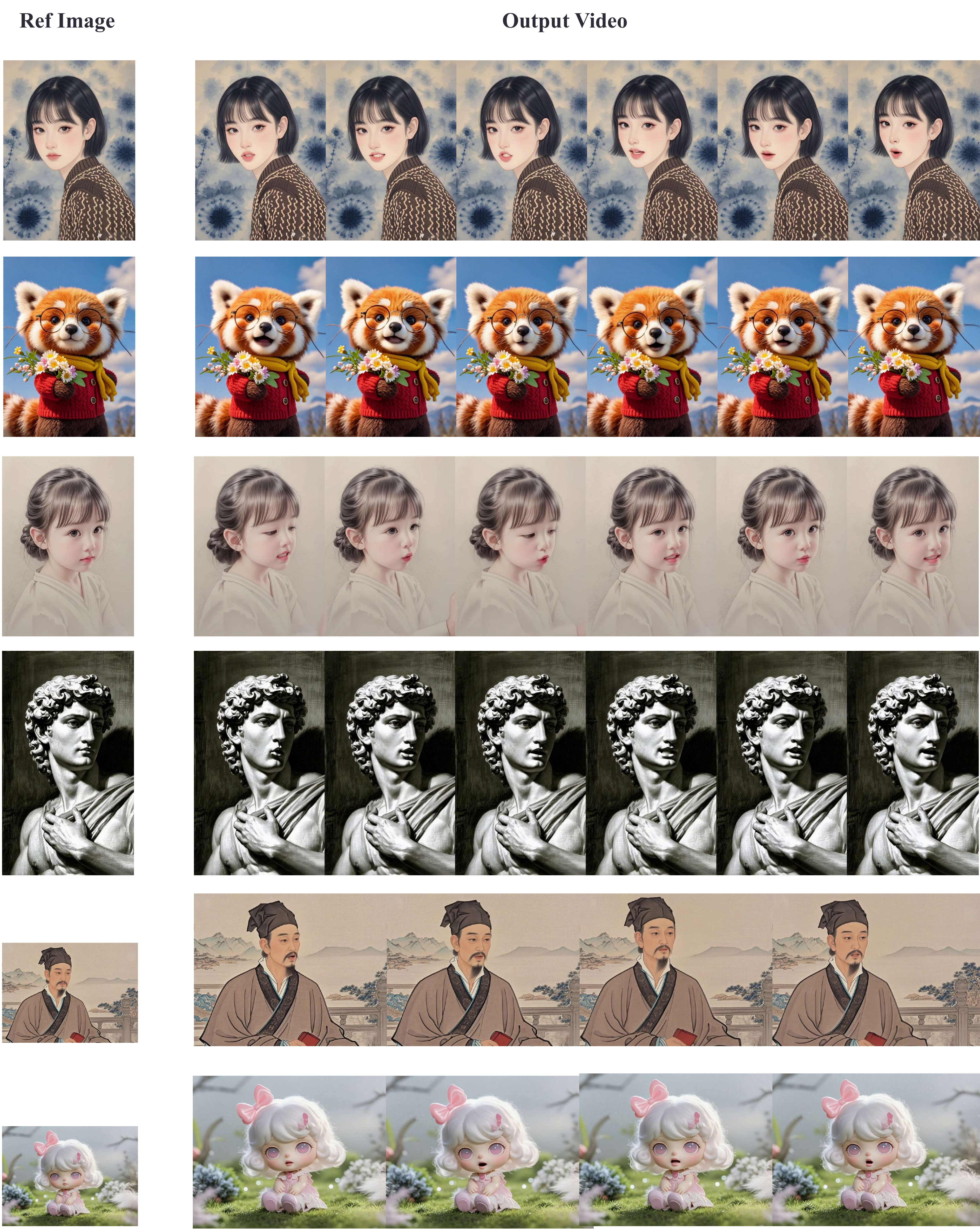}
    \caption{More visualizations on diverse character styles}
    \label{fig:supp3}
\end{figure}

Figure~\ref{fig:supp3} showcases the generation results of our method applied to characters with diverse styles. The results show that our method generalizes well across various styles, including LEGO, chinese painting, anime, and pencil sketch.

\begin{figure}[ht]
    \centering
    \includegraphics[width=1\textwidth]{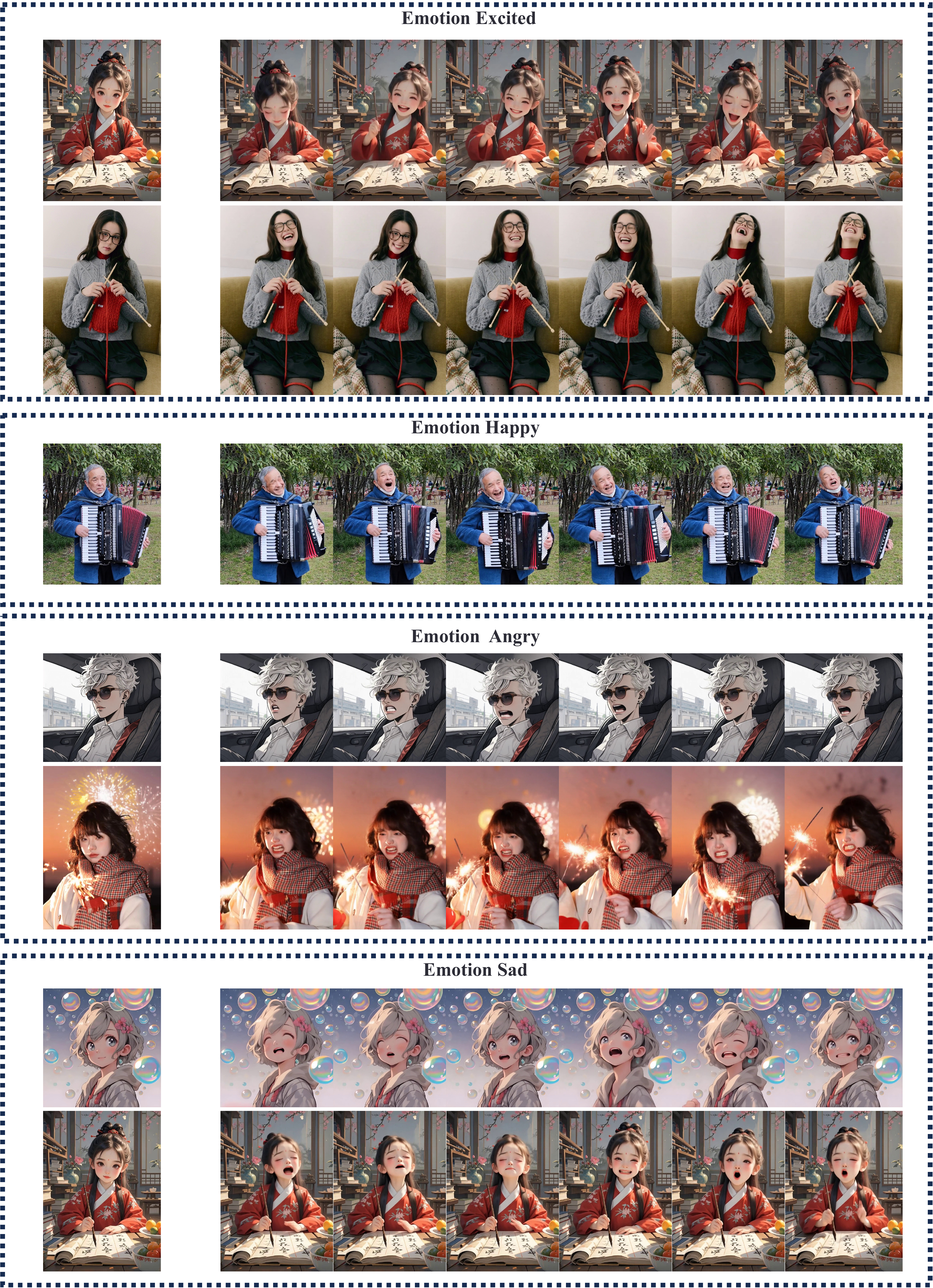}
    \caption{More visualizations on emotion control.}
    \label{fig:supp4}
\end{figure}

Figure~\ref{fig:supp4} demonstrates the precise control of emotions achieved by our method. It can be seen that our model has a good understanding of emotions such as happiness, sadness, excitement, and anger. This enables us to generate human animation videos that better align with the emotions conveyed by the audio, further demonstrating the unique capabilities of our model compared to previous audio-driven human animation methods.

In summary, compared to previous audio-driven human animation methods~\cite{wang2025fantasytalking, lin2025omnihuman, jiang2024loopy, hallo3, ji2024sonic}, our approach offers more practical features such as multi-character and emotion control audio-driven human animation. At the same time, it also outperforms previous methods in terms of character consistency and video dynamics. These advancements highlight the state-of-the-art performance and innovative design of our model.

\subsection{Limitations and Societal Impacts}
\label{sec:limitation}

\textbf{Limitations.} Firstly, our current approach relies on emotion reference images to drive the character's emotions, rather than allowing the model to infer and generate emotions directly from the audio. This leads to two main issues: (1) increased complexity for users during operation, and (2) the inability to reflect dynamic emotional changes within the video. Since each reference image corresponds to only one emotion, multiple emotions in a single audio segment may result in generation errors. Therefore, exploring methods to directly extract emotions from audio and generate corresponding emotional character videos is a promising direction for future research. Secondly, we currently use HunyuanVideo-13B~\cite{kong2024hunyuanvideo} as our base model, while FantasyTalking~\cite{wang2025fantasytalking} employs Wan2.1~\cite{wang2025wan}. Regardless of the base model, the inference process is time-consuming. For instance, generating a 10s video at 720×1216 resolution (with 50 inference steps) takes approximately 60 minutes, which is far from meeting the requirements of real-time applications. Thus, improving the model's generation speed to achieve real-time performance is one of our key future objectives. This will facilitate the application of our model in scenarios with higher real-time demands, such as live streaming and interactive real-time applications. 
Finally, exploring interactive human animation capable of real-time feedback is a promising research direction. This is expected to further expand the application of our method for users. This direction requires our model not only to possess strong content generation capabilities but also to have a solid understanding and contextual awareness, enabling fast and contextually appropriate responses to users.

\textbf{Societal Impacts.} Real-time interactive digital humans~\cite{ao2024body}  have become a major focus in the fields of artificial intelligence. However, their development has not yet reached its full potential due to several technical limitations. On one hand, current generative models still struggle to produce diverse and natural actions and expressions, making it difficult to achieve truly lifelike interactions. On the other hand, many high-performance models are extremely large in terms of parameter count, resulting in slow inference speeds that cannot meet the demands of real-time generation. These challenges significantly hinder the practical deployment of interactive digital humans.


\section{Contributors}
\definecolor{damaired}{RGB}{200, 0, 0}

\large{
\begin{itemize}
    \item \textbf{Project Leaders:} Qinglin Lu, Qin Lin, Yuan Zhou
    
    \item \textbf{Core Contributors:} Yi Chen, Sen Liang, Zixiang Zhou, Ziyao Huang, Yifeng Ma, Junshu Tang

    \item \textbf{Contributors:}
    Zhentao Yu, Zhengguang Zhou, Teng Hu, Zhiyao Sun, Yubin Zeng, Junxin Huang, Zhaokang Chen, Bin Wu, Xu Chen, Junwei Zhu, Chengjie Wang, Yuang Zhang, Junqi Cheng, Jiaxi Gu, Fangyuan Zou
\end{itemize}
}

